\documentclass[lettersize,journal]{IEEEtran}
\usepackage{amsmath,amsfonts}
\usepackage{algorithm}
\usepackage{array}
\usepackage[caption=false,font=normalsize,labelfont=sf,textfont=sf]{subfig}
\usepackage{textcomp}
\usepackage{stfloats}
\usepackage{url}
\usepackage{verbatim}
\usepackage{graphicx}
\usepackage{cite}

\usepackage{multirow}%
\usepackage{amssymb}%
\usepackage{amsthm}%
\usepackage{mathrsfs}%
\usepackage{xcolor}%
\usepackage{manyfoot}%
\usepackage{booktabs}%
\usepackage{algorithmicx}%
\usepackage{algpseudocode}%
\usepackage{listings}%
\usepackage[utf8]{inputenc} 
\usepackage{lineno}
\usepackage{mathtools}
\usepackage{svg}
\usepackage{pdfpages}
\usepackage{calc}
\usepackage{lmodern} 
\usepackage{soul}
\usepackage{array}

\algnewcommand{\algorithmicforeach}{\textbf{for each}}
\algdef{SE}[FOR]{ForEach}{EndForEach}[1]
  {\algorithmicforeach\ #1\ \algorithmicdo}
  {\algorithmicend\ \algorithmicforeach}

\newcommand{\swingPhase}{SW phase}
\newcommand{\stancePhase}{ST phase}

\begin{document}

\title{Model-based optimisation for the personalisation of robot-assisted gait training}

\author{Andreas Christou, Daniel F. N. Gordon, Theodoros Stouraitis, Juan C. Moreno and Sethu Vijayakumar
\thanks{This research was supported in part by the Engineering and Physical Sciences Research Council (EPSRC, grant reference EP/L016834/1) as part of the Centre for Doctoral Training in Robotics and Autonomous Systems at Heriot-Watt University and The University of Edinburgh, in part by the Alan Turing Institute, U.K., in part by Project I+D+i RED2022-134319-T (Spain), and in part by the Japan Science and Technology Agency (JST) Moonshot R\&D Program (Grant No. JPMJMS2239).}
\thanks{This paper has supplementary downloadable material available at
http://ieeexplore.ieee.org, provided by the authors. This includes one multimedia
MP4 format movie clip, which provides scenes of the experimental setup. This material is 24.1 MB in size.} \par
\thanks{A. Christou is with the School of Informatics, University of Edinburgh, EH8 9BT UK, (email: andreas.christou@ed.ac.uk). \par
D. F. N. Gordon is with Huawei's German Research Centre, 80992 Germany (email: daniel.gordon@huawei.com). \par
T. Stouraitis is with DeepSea Technologies, 105 64 Athens, Greece (email: stoutheo@gmail.com). \par
J. C. Moreno is with Centre of Automation and Robotics, 28500 Arganda, Spain (email: jc.moreno@csic.es). \par
S. Vijayakumar is with the School of Informatics, University of Edinburgh, EH8 9BT UK, (email: sethu.vijayakumar@ed.ac.uk)}%
\thanks{Manuscript received June, 2024}}

\markboth{Paper ID: TMRB-06-24-OA-0958}%
{Shell \MakeLowercase{\textit{et al.}}: A Sample Article Using IEEEtran.cls for IEEE Journals}

\maketitle

\begin{abstract}
Personalised rehabilitation can be key to promoting gait independence and quality of life. Robots can enhance therapy by systematically delivering support in gait training, but often use one-size-fits-all control methods, which can be suboptimal. Here, we describe a model-based optimisation method for designing and fine-tuning personalised robotic controllers. As a case study, we formulate the objective of providing assistance as needed as an optimisation problem, and we demonstrate how musculoskeletal modelling can be used to develop personalised interventions. Eighteen healthy participants (age = 26 $\pm$ 4) were recruited and the personalised control parameters for each were obtained to provide assistance as needed during a unilateral tracking task. A comparison was carried out between the personalised controller and the non-personalised controller. In simulation, a significant improvement was predicted when the personalised parameters were used. Experimentally, responses varied: six subjects showed significant improvements with the personalised parameters, \textbf{eight} subjects showed no obvious change, while \textbf{four} subjects performed worse. High interpersonal and intra-personal variability was observed with both controllers. This study highlights the importance of personalised control in robot-assisted gait training, and the need for a better estimation of human-robot interaction and human behaviour to realise the benefits of model-based optimisation.
\end{abstract}


\section{Introduction}\label{sec:intro}
Motor function deficits are often the result of neurological disorders and can significantly impact the quality of life of the affected person. It is well known that intensive multi-modal interventions can enhance the outcomes of physical therapy and help mitigate some of these deficits \cite{French2016,Yang2012}. However, in the case of gait rehabilitation, this can be a labour-intensive process, often involving several physical therapists to support a single patient. With the use of robotics, the physical strain on healthcare professionals can be alleviated and a means of delivering a systematic intervention can be provided. On the other hand, the use of robotic assistance in rehabilitation involves the risk of automating the process of rehabilitation and losing the highly personalised and effective treatment offered by physicians \cite{Pennycott2012d,Israel2007c}. As a result, there is currently a significant emphasis on employing collaborative robots to address the specific needs of patients and provide personalised assistance.
\begin{figure*}[h]
    \centering
    \includegraphics[width=0.99\textwidth]{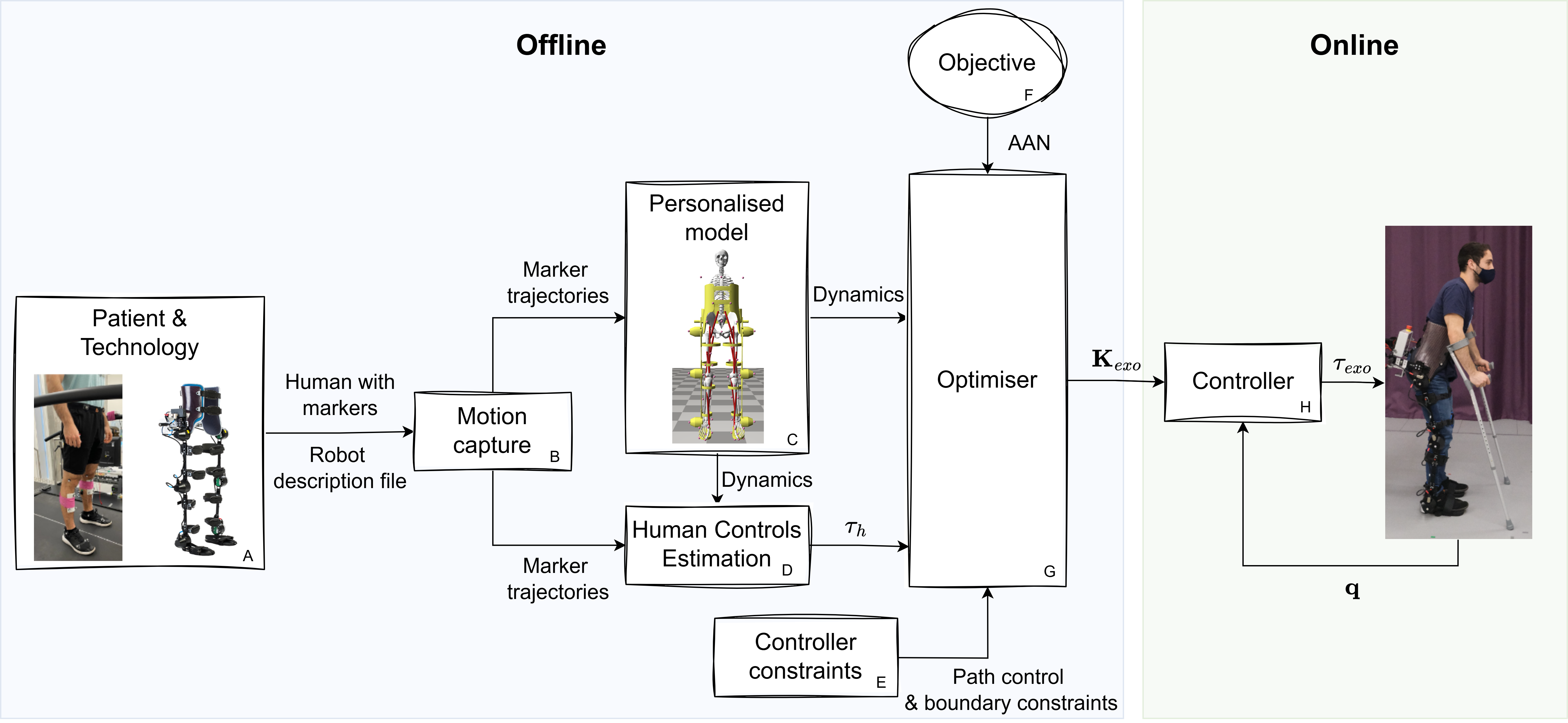}
    \caption{The offline model-based pipeline for control personalisation: (A) The physical properties and the motion of both the user and the robot are measured using (B) motion capture technologies to (C) create a personalised human-robot model and (D) obtain an estimate of the controls of the human for the completion of a task. (E) Together with the controller constraints and (F) the objectives of the collaborative task, (G) the optimal controller structure and/or controller parameters are obtained offline. (H) The outputs obtained from the offline optimisation are then used to design and/or fine tune the real-life robot controller to provide personalised assistance.}
    \label{fig:optimisation_pipeline}
\end{figure*}

For effective collaboration between human and robot in gait rehabilitation, the importance of providing partial assistance and encouraging the active participation of the patient has been highlighted in several studies \cite{Pennycott2012d,Kaelin-Lane2005,Lotze2003
}. This approach is often referred to as providing `assistance as needed' (AAN). However, even though there are several control strategies that can be used to provide assistance as needed \cite{Shahriari2019,Taherifar2018a,Hussain2017a,Russi2016c,Duschau-Wicke2010a}, it is unclear whether there is a single control strategy that can more effectively improve the functional outcomes of rehabilitation compared to other strategies, and that is in its form optimal for each individual. It is common, during the development of these strategies, to rely on a healthy participant's interaction with the robot in order to tune a controller that is intended for a group of people. As a result, most of these controllers have not been tailored to the specific needs of the individual and their effect has not been examined at the level of the individual. Given the high variability in gait among people, the same controller may not be optimal for everyone. It is therefore prudent to look at methods that can be used to adjust these controllers to meet the needs of each individual. 

The use of adaptive controllers has been proposed as a means of providing personalised assistance. These controllers are often based on impedance control, where a set of the controller's parameters, usually the controller gains, are adjusted on the fly according to the user's ability to complete the desired task \cite{Nasiri2021,Rajasekaran2018}. This added adaptive component to the control of rehabilitation robots appears to be promising, as it seems capable of providing assistance as needed \cite{Maggioni2018} and potentially improving unassisted gait \cite{Fleerkotte2014}. However, additional control parameters that need to be tuned are introduced, such as the initial conditions and the adaptive gains, which once again raises the question of how do we select these parameters to optimise assistance?

Another approach that is being studied for the personalisation of robotic controllers is human-in-the-loop (HIL) optimisation \cite{Gordon2022, Ding2018, Zhang2017}. HIL optimisation allows for the online optimisation of different control parameters in order to minimise the value of an objective function. Commonly, this objective function is optimised iteratively, using `derivative-free' optimisation techniques, such as Bayesian optimisation or the covariance matrix adaptation evolution strategy (CMA-ES), during the execution of the desired task and while the user is wearing the robot. However, this approach has mostly been applied for the purpose of reducing the metabolic cost of either unimpaired subjects \cite{Gordon2022, Ding2018, Zhang2017} or amputees \cite{Welker2021}. Only recently has the application of HIL optimisation been explored in the field of rehabilitation \cite{Wang2020}. Wang et al. \cite{Wang2020} used HIL optimisation to adjust the difficulty of a cycling game. In their study, Wang et al. \cite{Wang2020} showed that by adjusting the reference speed of the game using HIL optimisation, the participants were able to more accurately track the desired speed while maintaining high muscle activation. While more studies are needed to evaluate the effectiveness of HIL optimisation in robot-assisted rehabilitation, one limitation associated with this process is the time-consuming nature of the online iterative search for the global optimum.

Here, we describe an offline model-based approach for the design and personalisation of assistive controllers (Figure \ref{fig:optimisation_pipeline}), where we leverage the power and advances in musculoskeletal modelling, which has been recognised as a powerful tool for both understanding the physical properties of biological systems, and informing the design of personalised devices and interventions \cite{Fregly2021,Quinn2023}. We hypothesise that by observing human motion and the human-robot interaction through the lens of motion capture systems, we can build personalised human-robot models and develop individualised assistive controllers through offline optimisation in order to improve the collaboration between humans and robots without the need for extensive human-in-the-loop experiments. In this manuscript, we firstly present our model-based framework in a generalised form to highlight its utility in similar fields where the personalisation of human-robot interaction is important, such as human augmentation and/or (industrial) human-robot collaboration. Next, we demonstrate the ability to personalise an AAN controller of a lower-limb exoskeleton to the needs of the user as a case study. We formulate the concept of providing assistance as needed as an optimisation problem, and we illustrate how we can obtain a personalised controller for each individual. With the help of eighteen healthy participants, we carry out the comparison between the personalised controller and the non-personalised controller, and evaluate the effect of our proposed approach on the collaboration between the participants and the robot. The key contributions of this work involve, firstly, the introduction of an innovative methodology for the personalisation of robotic controllers in human-robot collaboration utilising musculoskeletal modelling, and secondly, the evaluation of this methodology, specifically for personalising an AAN controller, through testing on healthy individuals.

\section{Methods}
This section describes the different components of our proposed method as illustrated in Figure \ref{fig:optimisation_pipeline}. The generic personalisation framework is described first, followed by a description of the human-robot model, the estimation of the human controls, and the definition of the controller objectives and constraints used to fine tune the robot controller \cite{Christou2023}.  

\subsection{Controller optimisation framework}
The aim of our model-based optimisation framework is to minimise a set of (rehabilitation) costs, $C$, given a model of the combined human-robot system, $S$, in order to obtain a set of personalised controller parameters, $P^{*}$. This can be expressed as a multi-objective optimisation problem in the following generalised form: 

\begin{align}
\min_{\bf P} {\sum_{i=1}^{N} w_{i}C_{i}(S)} \label{eq:ObjFunGeneric} \\
g({\bf x},{\bf u}) = 0,  \label{eq:generalisedConstraints1}\\
h({\bf x},{\bf u}) > 0, \label{eq:generalisedConstraints2} \\
{\bf x}^{-}<{\bf x}< {\bf x}^{+}, \label{eq:generalisedConstraintsEnd-1} \\
{\bf u}^{-}<{\bf u}< {\bf u}^{+}, \label{eq:generalisedConstraintsEnd}
\end{align}
where $g(\cdot)=0$ and $h(\cdot)>0$ represent the generalised equality and inequality constraints, respectively (the specifics of which will be discussed in more detail in the following sections), and $w$ represents the weight of the associated cost terms, for $N$ number of costs. $\bf u$ is the vector of controls of the human model and the robot model, and ${\bf x}$ is the vector of states of the human-robot model, such as the joint positions and velocities. $[{\bf x}^{-}, {\bf x}^{+}]$ and $[{\bf u}^{-}, {\bf u}^{+}]$ are the lower and upper bounds of the model's states and controls, respectively.

\subsection{Human-robot model} \label{sec:human-robot-model}
With the use of musculoskeletal modelling software, such as OpenSim, a personalised human-robot model can be constructed \cite{Delp2007}. OpenSim offers detailed musculoskeletal models with variable levels of complexity for the lower limbs, the upper limbs, the back or the full body, that can be adjusted to reflect the physical properties of the human, including any characteristics that may be a result of an injury, physical conditioning or a neurological disorder. This adjusted human model can then be combined with the robotic description file of the desired robot in order to create a personalised human-robot model (Figure \ref{fig:optimisation_pipeline}C). The coupled dynamics of this model are a fundamental constraint of the optimisation problem (Equation \ref{eq:generalisedConstraints1}) and can be expressed as:
\begin{align}
    {\bf M}_{hr}{(\bf q)}{\bf \ddot{q}} + {\bf C}_{hr}({\bf q},{\bf \dot{q}}) +  &{\bf G}_{hr}({\bf q}) = {\boldsymbol \tau}_{h} \notag\\ &+ {\boldsymbol \tau}_{r} + {\bf J}({\bf q})^{T}{\bf f}_{\text{ext}}, \label{eq:dynamic_model}
\end{align}
where ${\bf q}, {\bf \dot{q}}, {\bf \ddot{q}} \in \mathbb{R}^{n}$ are the generalised joint positions, velocities and accelerations of the model, respectively. ${\bf M}_{hr}{(\bf q)} \in \mathbb{R}^{n \times n}$ is the mass matrix of the human-robot model, ${\bf C}_{hr}({\bf q},{\bf \dot{q}}) \in \mathbb{R}^{n}$ is the vector of Coriolis and centrifugal forces, and ${\bf G}_{hr}({\bf q}) \in \mathbb{R}^{n}$ is the vector of gravitational forces for a system with $n$ degrees of freedom. ${\boldsymbol \tau}_{h} \in \mathbb{R}^{n}$ represents the human's voluntarily generated joint torques, and ${\boldsymbol{\tau}}_{r} \in \mathbb{R}^{n}$ are the assistive forces provided by the robotic device. ${\bf J} \in \mathbb{R}^{3 \times n} $ is the system's Jacobian and ${\bf f}_{ext} \in \mathbb{R}^{3}$ represents any external forces that may be applied to either the robot or the human model, including forces due to the human-robot interaction and ground reaction forces.

For this study, the generic OpenSim model, \textit{gait1018}, was used. This human model focuses on the lower limbs and consists of 10 degrees of freedom and 18 musculotendon units. To construct personalised human-exoskeleton models, \textit{gait1018} was first scaled using motion capture technology. 

\begin{figure}[h]
    \centering
    \includegraphics[width=0.99\linewidth]{{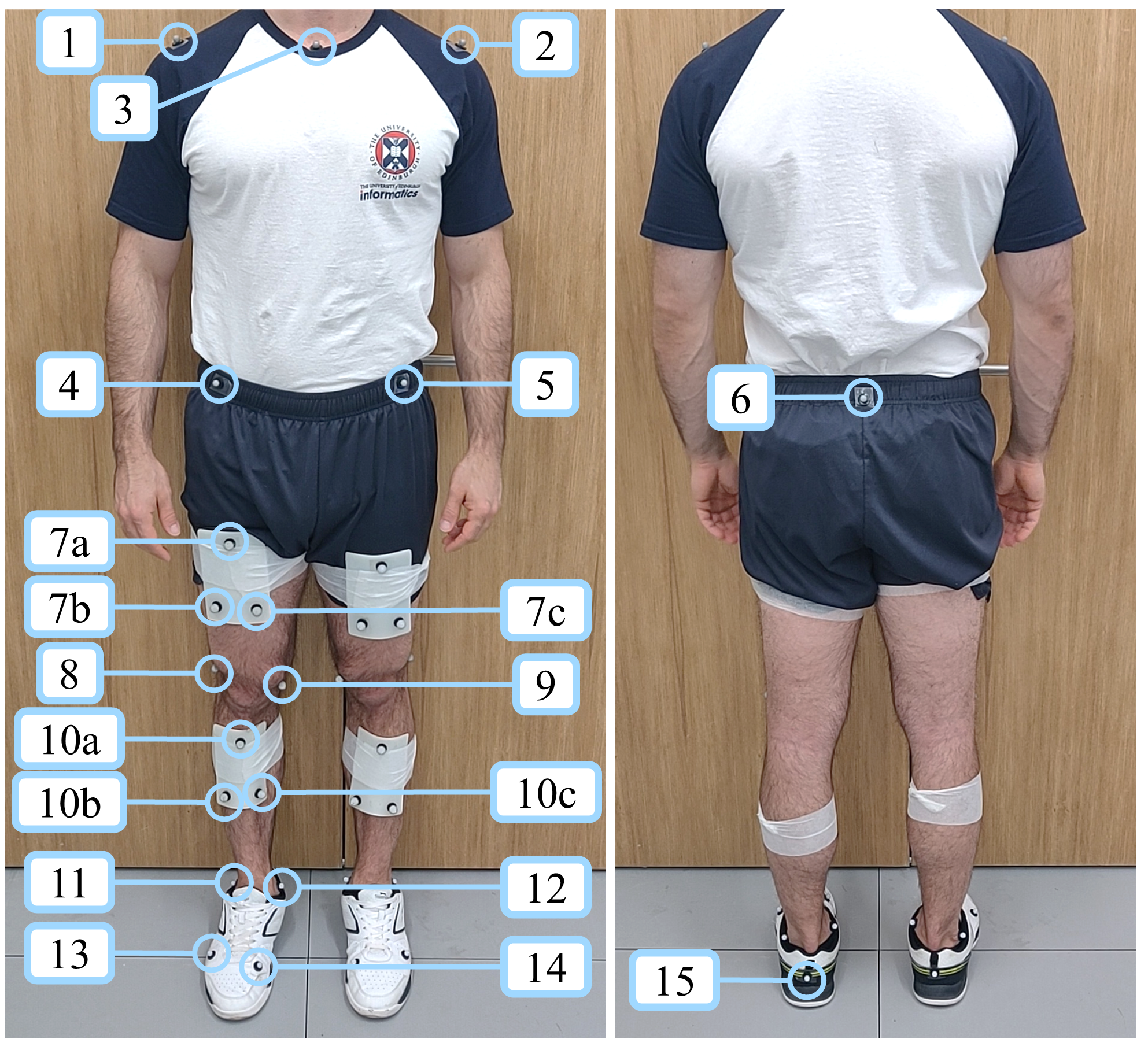}}
    \caption{Placement of reflective markers for model scaling. Tracking markers were placed on the right (1) and the left (2) shoulder and an anatomical marker was placed on the sternum (3). Three anatomical markers were used to define the pelvis with one marker on the right anterior superior iliac spine (ASIS) (4), one on the left ASIS (5) and one on the sacrum (6). A cluster of three tracking markers was used on the thighs (7a-7c) and anatomical markers were placed on the lateral (8) and medial (9) femoral epicondyles. A cluster of three tracking markers was also placed on the shanks (10a-10c). The feet were defined with five anatomical markers placed on the lateral (11) and medial (12) malleoli, the fifth (13) and the first (14) metatarsal heads and the heel (15).}
    \label{fig:vicon_markers}
\end{figure}

For the scaling of the human model reflective markers were placed on the participants and a static pose was recorded using a 12-camera Vicon system. Anatomical marker pairs were defined for each segment, and OpenSim's scaling tool was used to scale the model's geometry and inertial properties (Figure \ref{fig:vicon_markers}). A marker adjustment was also carried out in OpenSim in order to minimise the error between the model markers and the experimental markers. The accuracy of the scaled model was evaluated using OpenSim's GUI, where the model's joint coordinates were reviewed, as well as the mean and maximum RMS error achieved by the marker adjustment process.

Onto the scaled human model, the dynamic model of the exoskeleton Exo-H3 was added (Technaid, Spain). This involved the adjustment of the exoskeleton's configuration and the modelling of the interaction forces between the human and the robot. Firstly, the exoskeleton's orientation and position in the medial-lateral plane was adjusted, and the hip joint centre of the exoskeleton was aligned with the hip joint centre of the human model in the sagittal plane. With the hip joints aligned, the length of the exoskeleton's limbs was adjusted sequentially from top to bottom in order to achieve a good alignment between the joint centres of the two models. This exoskeleton adjustment was performed according to the constraints imposed by the hardware, and the fitness of the human-robot model was evaluated using OpenSim's GUI.

For the modelling of the interaction forces between the human and the robot, linear bushing forces were used at the locations of the exoskeleton's straps. These are translational and rotational triaxial springs and dampers that were placed at the torso, the thighs (lower and upper thigh strap), the shanks (lower and upper shank strap), and the feet (heel contact and foot strap). The stiffness of the bushing forces was selected empirically using OpenSim's GUI to ensure a firm contact between the exoskeleton model and the human model. The values of the bushing forces chosen for the different contact points are presented in Table \ref{table:bushing forces}.
\newcolumntype{P}[1]{>{\centering\arraybackslash}p{#1}}
\begin{table}[h]
    \centering
    \caption{Stiffness of translational (x, y, z) and rotational ($\theta, \phi, \psi$) bushing forces for the modelling of human-robot contact, where K and B are the stiffness and damping coefficients, respectively.}
    \begin{tabular}{c c c}
    \toprule
    & Torso & Thighs, Shanks \& Feet \\
    \midrule
    \rule{0pt}{3ex} 
    \begin{tabular}[c]{@{}c@{}}{[$K_{x}$ $K_{y}$  $K_{z}$]}\\ { (kN/m)}\end{tabular}  &   [40 40 40]    &     [10 10 10]  \\[2ex]
    \rule{0pt}{3ex} 
    \begin{tabular}[c]{@{}c@{}}{[$B_{x}$ $B_{y}$ $B_{z}$]}\\ {(kN/m/s)}\end{tabular}  &    [0.2 0.2 0.2]   &     [0.1 0.1 0.1]  \\[2ex]
    \rule{0pt}{3ex} 
    \begin{tabular}[c]{@{}c@{}}{[$K_{\theta}$ $K_{\phi}$ $K_{\psi}$]}\\ {(kN/rad)}\end{tabular}    &    [1 1 1]   &    [0.1 0.1 0.1]  \\[2ex]
    \rule{0pt}{3ex} 
    \begin{tabular}[c]{@{}c@{}}{[$B_{\theta}$ $B_{\phi}$ $B_{\psi}$]}\\ {(kN/rad/s)}\end{tabular}  & [0.03 0.03 0.03] & [0.01 0.01 0.01]  \\[2ex]
    \bottomrule
    \end{tabular}
\label{table:bushing forces}
\end{table}

Lastly, to reduce the complexity of the optimisation problem and the computational demands of this process, the model's muscles were replaced by ideal joint actuators. The upper and lower limits of the joint actuators were defined based on the values reported in \cite{Sarabon2021,Krantz2020,Moraux2013}.

\subsection{Estimation of human controls} \label{sec:human behaviour model}
In the context of human-robot collaboration, human behaviour becomes a primary contributor to task completion. Contrary to other scenarios of human-robot interaction where human behaviour may be considered a disturbance, here it is increasingly important to recognise it as a central element in the collaborative dynamics. Therefore, for the model-based personalisation of the robot controller, a reliable model of the human behaviour, ${\boldsymbol \tau}_{h}$, is required. In many cases, human behaviour has been characterised by a combination of feedforward and feedback processes with both long-term and short-term adaptations, where feedforward processes become predominant as adaptation progresses \cite{Maeda2020,Pisotta2014,Haith2013,Seidler2004}. Computational models that describe the adaptation of human motor control in environments with high uncertainty and environments with predictable external perturbations have been proposed in \cite{Tee2010,Scheidt2001,Thoroudhman2000}, while efforts have been made to also capture human behaviour during gait and sit-to-stand using inverse optimal control \cite{Gordon2023,Hoa2021,Nguyen2019,Geravand2017}. However, using inverse optimal control is very computationally expensive, and the computational models presented in \cite{Tee2010,Scheidt2001,Thoroudhman2000} have mostly focused on upper-limb end-effector motion in one dimension and in environments with perturbing force fields, which may not translate to the application of assistive forces in a two-dimensional joint space for the lower limbs.

Therefore, for this study a feedforward model was used (Figure \ref{fig:optimisation_pipeline}D). To do this, the motion of the human was recorded while performing the desired task (see section \ref{sec:experimental_setup}), and while wearing the exoskeleton in transparent mode, (${\boldsymbol \tau}_{r}=0$). Five consecutive cycles were extracted from the recorded data and a PD controller was used at the joints of the human model, in a forward dynamics analysis, to estimate the human joint torques that are necessary to reproduce, in simulation, the recorded motion. This process was carried out using the Residual Reduction Algorithm (RRA) available in OpenSim. The obtained joint torques, ${\boldsymbol \tau}_{h}$, were then used as a constraint in the optimisation pipeline to replace this human PD controller (Equation \ref{eq:generalisedConstraintsEnd}), and form the feedforward human model. To prevent any biases introduced from the short-term adaptation of the human and/or the initiation and the termination of the movement, a training period was prescribed prior to the data collection, and cycles from the start and the end of the recording were excluded. Due to the short duration of the study, it was assumed that adaptation beyond the training phase will be insignificant.

\subsection{Robot control model}
Given the human-robot model and a model for the human behaviour, it is now possible to identify in simulation the optimal robot behaviour, ${\boldsymbol \tau}_{r}$. For this, a robot control model with the respective constraints can be defined (Figure \ref{fig:optimisation_pipeline}E). This involves constraints regarding the controller's structure and/or the controller's limits (Equations \ref{eq:generalisedConstraints1}-\ref{eq:generalisedConstraintsEnd}). Unlike the human-robot model and the human behaviour model, these constraints are optional, since it is possible to use unconstrained optimisation from which the optimal controller structure or appropriate limits may be inferred.

\begin{figure}[h]
    \centering
    \includegraphics[width=0.99\linewidth]{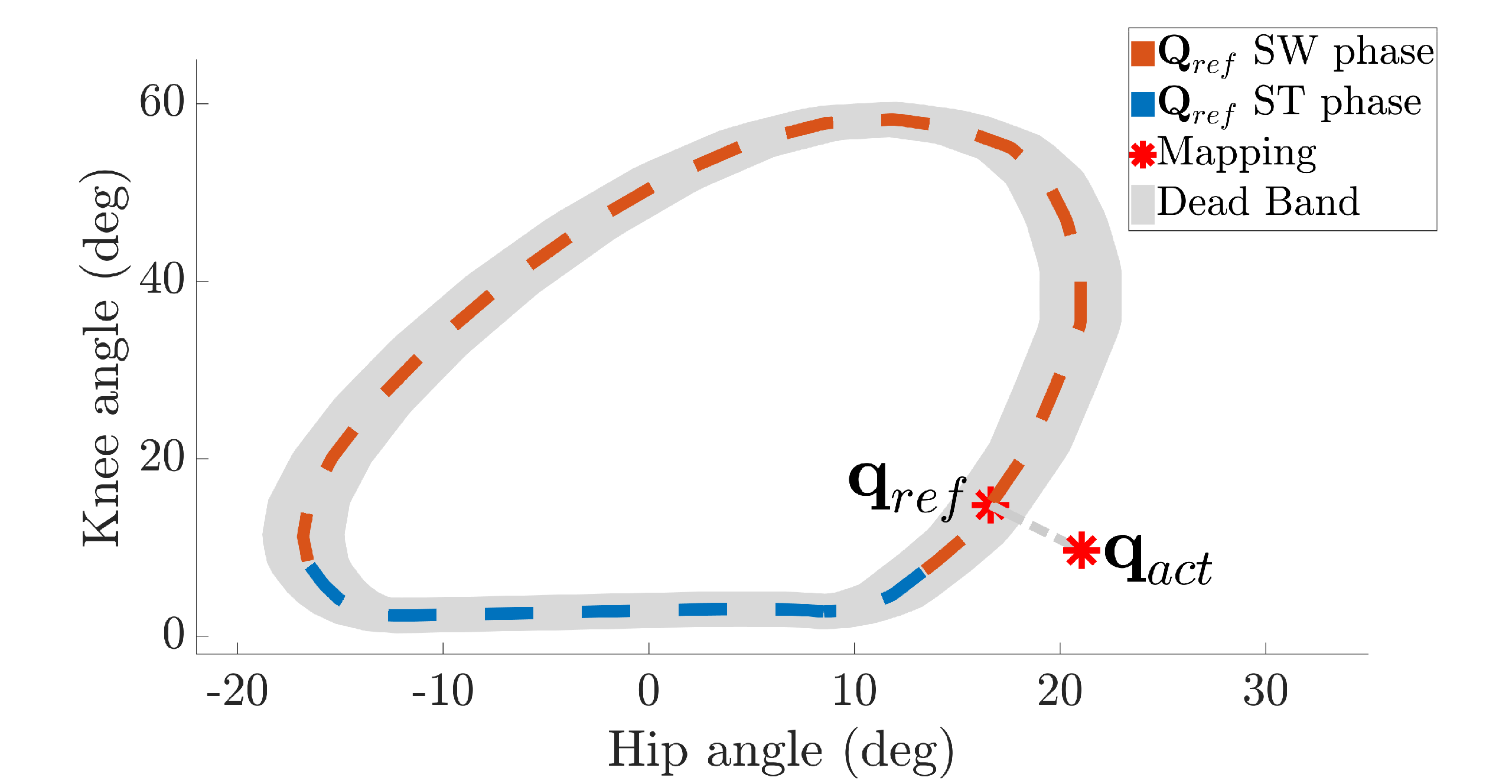}
    \caption{Illustration of the dual-phase reference kinematic path, ${\bf Q}_{ref}$, defined in joint space, and an instance of the dynamic allocation of the reference point, ${\bf q}_{ref}$, based on the pose of the exoskeleton, ${\bf q}_{act}$.}
    \label{fig:path_control_ref_act_mapping}
\end{figure}

For this study, a case is demonstrated where both the controller structure and the controller limits are provided. This involves the personalisation of an impedance controller, which is based on path control \cite{Duschau-Wicke2010a,Duschau-Wicke2010b}. Path control uses a reference kinematic path in joint space, ${\bf Q}_{ref} \in \mathbb{R}^{i \times 2}$, that describes the desired relationship between the hip joint angle and the knee joint angle in the sagittal plane (where $i$ is the number of points in the discretised domain of the reference path). Based on this reference path, the reference point, ${\bf q}_{\text{ref}} \in \mathbb{R}^{2}$, is dynamically defined as the point on the reference path where the Euclidean distance between the reference path, ${\bf Q}_{ref}$, and the joint angles of the human (herein referred to as the actual point), ${\bf q}_{act} \in \mathbb{R}^{2}$, is at a minimum (Figure \ref{fig:path_control_ref_act_mapping}). 

The absolute joint angle error, ${\Delta \bf \tilde q} \in \mathbb{R}^{2}$, is then defined as the difference between the reference point and the actual point. In order to allow for some error tolerance, a dead band region of radius, $r_{db}$, is defined around the reference path and the true joint angle error, ${\Delta {\bf q}} \in \mathbb{R}^{j}$, is defined as the difference between the reference point and the actual point, minus the error tolerance when the error tolerance is exceeded. This is expressed as \cite{Duschau-Wicke2010b}: 
\begin{gather}
    {\Delta{\bf \tilde q}} = {{\bf q}}_{\text {ref}} - {\bf q}_{\text {act}}, \label{eq:pathControl1} \\
    {\Delta {q}}^{(j)} =
    \begin{cases}
        0 , & |{\Delta \tilde q}^{(j)}| \leq r_{\text {db}}, \\
        {\Delta \tilde q}^{(j)} - r_{\text {db}} , & {\Delta \tilde q}^{(j)} > r_{\text {db}},  \\
        {\Delta \tilde q}^{(j)} + r_{\text {db}} , & {\Delta \tilde q}^{(j)} < -r_{\text {db}}.
    \end{cases}
\end{gather}

Based on the true joint angle error, ${\Delta {\bf q}}$, a proportional-derivative (PD) controller is used to ensure that assistive forces are provided by the exoskeleton in order to guide the user closer to the reference path. This is expressed as:
\begin{gather}
    {\boldsymbol \tau}_{exo} = {\bf K}{\Delta{{\bf q}}} + {\bf B}{\Delta{{\dot{\bf q}}}}, \label{eq:impedanceController1} \\
    {\bf B}={\bf c}_{cr}\sqrt{\bf K}, \label{eq:impedanceController2}
\end{gather}
where $\bf K$ and $\bf B$ are the joint stiffness and damping matrices of the exoskeleton's joints, and ${\bf c}_{cr}$ is the matrix of the critical damping coefficients. 

For this study, we defined the reference path using the kinematic data collected from a healthy subject during walking (S2). This reference path was adapted to a path with a less pronounced loading response, and a dead band with a radius of 2 degrees was defined around it. Based on the recorded data, the reference path was divided into two phases (as shown in Figure \ref{fig:path_control_ref_act_mapping}): the stance phase (ST), and the swing phase (SW). A PD controller with constant stiffness was then defined for each phase: 
\begin{gather}
    {\bf K}=
    \begin{cases}
    {\bf K}_{st}, & \text{\stancePhase} \\
    {\bf K}_{sw}, & \text{\swingPhase} \\
    \end{cases} \label{eq:impedanceController3}
\end{gather}

As a result, the open parameters for the given controller are the stiffness of the exoskeleton's hip joint and knee joint during the two cycle phases. Thus, the decision variables, $\bf P$, of the optimisation problem (Equation \ref{eq:ObjFunGeneric}), consist of four parameters: the hip stiffness, $K_{hsw}$, and the knee stiffness, $K_{ksw}$, during SW, and the hip stiffness, $K_{hst}$, and the knee stiffness, $K_{kst}$, during ST.

\subsection{Offline optimisation}
Using the personalised human-robot model, the expected human joint torques, and the constraints implemented due to the exoskeleton controller (Equations \ref{eq:pathControl1}-\ref{eq:impedanceController3}), forward dynamics simulations can be carried out to obtain the exoskeleton behaviour that minimises the desired cost. Using forward dynamics, the kinematics of the human-exoskeleton model can be predicted for different levels of assistance as follows:
\begin{align}
    {\bf \ddot{q}} =  {\bf M}_{hr}^{-1}{({\bf q})}(&{\boldsymbol \tau}_{h} +{\boldsymbol \tau}_{r} + {\bf J}({\bf q})^{T}{\bf f}_{\text{ext}} \notag\\ &-{\bf C}_{hr}({\bf q},{\bf \dot{q}}) - {\bf G}_{hr}({\bf q})).   
    \label{eq:dynamic_model_rearranged_FD} 
\end{align}
Based on this predicted trajectory, the objective function value can be obtained (Equation \ref{eq:ObjFunGeneric}). 

For this study, we used an objective function that sums up the true joint angle error, ${\Delta \bf q}$, and the exoskeleton controls,  ${\bf u}_{r}$, throughout the predicted motion (Figure \ref{fig:optimisation_pipeline}F). The selection of these two terms in the objective function is based on the concept of providing assistance as needed in robot-assisted rehabilitation. This implies that the robotic device is expected to guide the patient towards the desired direction but with minimal assistive forces, such that the patient is encouraged to use their residual strength. This can be expressed as:
\begin{align}
\min_{\bf K} \quad \frac{w_{1}}{J_{1}}\frac{{\sum_{i=1}^{N-1} {\bf u}_{{r}_{i}}^{T}{\bf I}{\bf u}_{{r}_{i}}}}{N-1} + \frac{w_{2}}{J_{2}}\frac{\sum_{i=1}^{N}{{\Delta \bf q}_{i}^{T}}{\bf I}{{\Delta \bf q}_{i}}}{N},
\label{eq:ObjFun}
\end{align}
where $w_1$ and $w_2$ are the weights of the two costs, $\bf I$ is the identity matrix, $N$ is the number of the simulation time steps and $J$ is a scaling factor. The scaling factor is used to normalise the cost terms to the maximum exoskeleton assistance and the maximum expected error, respectively, such that the magnitude of the two costs is comparable. Similarly, $N$ is used to normalise the two cost terms to the length of the simulated cycles and $w$ is used to adjust the relative importance of the two normalised costs. 

To solve this problem using OpenSim, where the analytical form of the dynamics is not available, a gradient-free optimisation tool, \textit{surrogateopt}, was used (Figure \ref{fig:optimisation_pipeline}G). \textit{Surrogateopt} is a global solver available in MATLAB, which is designed for computationally expensive functions that may not be smooth. Given finite bounds on the decision variables, \textit{surrogateopt} explores the unknown optimisation landscape, and uses cubic radial basis function interpolation to create a surrogate function \cite{Gutmann2001,Powell1992}. It balances exploration of the search space with exploitation of promising regions using a merit function \cite{wang2014}, and iteratively refines the surrogate model to improve its predictions. After the predefined number of iterations is exceeded, this process terminates, and the arguments of the minimum observed point are provided. For the given size of the optimal control problem, we empirically observed that convergence is typically reached within 50-70 iterations. Therefore, a total of 150 iterations was defined as the maximum number of iterations. 
The following pseudocode describes the way this optimisation process was carried out for the personalisation of the exoskeleton's stiffness:
\begin{algorithm}
\caption{Pseudocode for offline controller optimisation}\label{alg:cap}
\begin{algorithmic}[1]
\Require $0 < {\bf w} < 1$
\State $N \gets 150$; $i \gets 1$; $\bf S \gets \{\}$
\State $[{\bf x}^{-}, {\bf x}^{+}] \gets [0, 600]$
\While {$i < N$}
\State ${\bf K}_{i} \gets \text{surrogateopt}(\bf S, {\bf x}^{-},{\bf x}^{+})$
\State ${\bf \ddot{q}} \gets \text{forward\_dynamics}({\boldsymbol \tau}_{h}, {\bf K}_{i}) $ 
\State ${\bf O}_{i} \gets \text{objective\_fcn\_value}({{\Delta \bf q}}, {\bf u}_{e}, {\bf w})$
\State ${\bf S} \gets \{{\bf S}; [{\bf K}_{i}, {\bf O}_{i}]\}$
\State $i \gets i+1$
\EndWhile
\State ${\bf K}^{*} \gets {\bf S}$ \text{such that } ${\bf O}^{*}=\min \bf S$
\end{algorithmic}
\end{algorithm}

\subsection{Online controller}
Once the offline optimisation is completed, the outputs can be used to tune the hardware's controller in order to provide personalised assistance (Figure \ref{fig:optimisation_pipeline}H). In the current case, where a specific controller structure is desired, this involves the implementation of a real-time feedback controller and the adjustment of the gains according to the optimisation outputs. 

\section{Evaluation with healthy participants}
\subsection{Subjects}
The effectiveness of this control personalisation approach was studied with the help of eighteen healthy volunteers (15 male, 3 female, age = 26 $\pm$ 4, weight = 74.6kg $\pm$ 10.4kg). All participants were first-time users of a wearable robot and had no prior knowledge of the task to be completed. The experiment pipeline was approved by the University of Edinburgh, School of Informatics Ethics Committee (ID 2021/46920) and all participants provided written consent.

\subsection{Hardware}
For the motion capture of the participants a 12-camera Vicon system was used, and the exoskeleton Exo-H3 was used to provide assistance during the task (an upgraded version of the Exo-H2 exoskeleton \cite{Bortole2015a}, Technaid, Spain). The exoskeleton controller was developed in Simulink Desktop Real-Time and operated at 100Hz. A separate Windows PC was used to provide real-time visual feedback to the participants during the training period. All simulations were carried out on a system equipped with an Intel Core i9-9900KF 3.60 GHz CPU, featuring 8 cores, 16 logical processors and 64 GB of RAM operating at 3200 MHz. The operating system was Windows 10 Pro, and the software environment included an interface between MATLAB and OpenSim. The optimisation process was parallelised by activating \textit{surrogateopt}'s parallel option, and each simulation, which processed approximately 2300 time steps for a total of 150 optimisation iterations, completed in approximately 20 minutes.  Computational demands were highly dependent on the complexity of the human-robot model, the stiffness of the interaction forces, and the number of integration steps required for each simulation. As a result, the duration of each participant's visit lasted approximately 2 hours.

\subsection{Experimental setup} \label{sec:experimental_setup}
\begin{figure*}[h]
    \centering
    \includegraphics[width=0.99\textwidth]{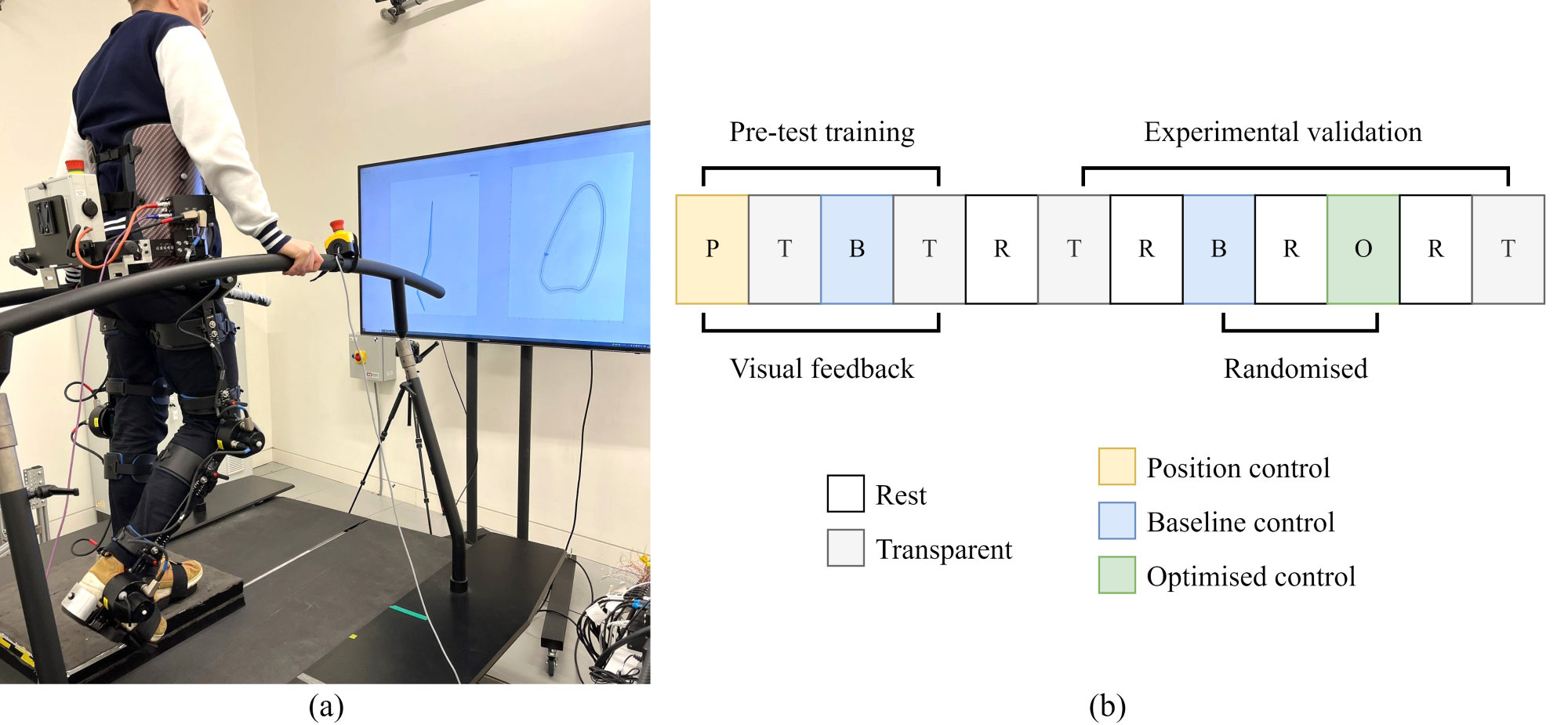}
    \caption{(a) Data collection setup. Participant wearing an exoskeleton and performing a unilateral tracking task with the help of visual feedback. (b) Experimental protocol including four phases of pretest training with visual feedback and the experimental validation where the baseline controller and the optimised controller were tested in a randomised order.}
\label{fig:experimental_protocol}
\end{figure*}

The participants were first fitted with reflective markers, as explained in section \ref{sec:human-robot-model}, to enable the modelling of the human-exoskeleton system. Once a scaled model was constructed, the markers were removed and the participants were fitted with the exoskeleton. The exoskeleton was adjusted to the dimensions of each participant to ensure a tight and comfortable fit, and a good alignment between the joints of the exoskeleton and the joints of the participant. While wearing the exoskeleton, the participants were asked to perform a trajectory tracking task with their right leg, while their left leg was used to support their weight on an elevated platform, which was used to avoid contact between the user's right leg and the ground (Figure \ref{fig:experimental_protocol}). Side rails were provided to help the participants maintain their balance. This task was selected in order to reduce the computational demands of the optimisation, and uncertainties involved with the optimisation of balance.

The participants were then informed about the experimental protocol and were asked to follow the steps shown in Figure \ref{fig:experimental_protocol}b. At all stages, the participants were informed about the state of the exoskeleton. Prior to recording the participants' performance, a training period was prescribed to allow the participants to get familiar with both the task and the exoskeleton (Figure \ref{fig:experimental_protocol}). This training period included four phases complemented by visual feedback: (1) a position controlled exoskeleton (${\bf q}_{exo}(t)={\bf q}_{ref}(t)$), (2) a transparent exoskeleton (${\boldsymbol \tau}_{e}=0$), (3) an assistive exoskeleton (${\boldsymbol \tau}_{e} \neq 0$), and (4) a transparent exoskeleton again. Each training phase lasted between 2-5 minutes, depending on the user's confidence level. The visual feedback included a virtual image of the exoskeleton's leg, as well as a graphical representation of the reference kinematic path in joint space with the real-time pose of the exoskeleton and the corresponding reference point (Figure \ref{fig:experimental_protocol}a). The visual queues were explained to the participants and the participants were instructed to follow the reference path as accurately as possible. After the training period, the visual feedback was removed for all experiments following the training period to prevent further learning of the task and capture the participant's kinematics that reflect the internal model of motor control constructed by the participants during this period. The participants were asked to perform the task they practiced with the exoskeleton in transparent mode and their motion was recorded and used for the estimation of the human behaviour model.

Based on this recorded motion, a personalised set of exoskeleton stiffnesses was obtained for each participant through our offline model-based optimisation method. This included the exoskeleton's stiffness for both the hip and the knee joints during the two different phases as shown in Figure \ref{fig:path_control_ref_act_mapping}. The obtained stiffness parameters were used to adjust the online exoskeleton controller, and the participants were asked to perform the desired task while wearing the exoskeleton in assistive mode. Their performance while using the exoskeleton with both their personalised stiffness and a baseline stiffness was recorded and analysed. In this case, the baseline stiffness was set to 340Nm/rad, which is in line with the stiffness used in \cite{Del-Ama2014a} and approximately half of the maximum stiffness used in \cite{Duschau-Wicke2010a}. The order at which the participants experienced the baseline controller and the optimised controller was randomised using MATLAB's random number generator, \textit{randi}. The metrics used to verify the effectiveness of the control personalisation included the kinematic tracking error of the participants, the level of assistance they received from the exoskeleton, and the weighted sum of the two, which formed the objective function of the optimisation problem (Equation \ref{eq:ObjFun}). At the end of the experiment, the participants were asked to perform the task while wearing the exoskeleton in transparent mode once again. This was done to assess any changes that may have occurred during the experiment in the participants' behaviour, and test the validity of our previous assumption that the participants' behaviour will not significantly change during the experiment beyond the training phase.

\subsection{Analysis}
Given the recorded motion of the participants, two-sided permutation tests with 100,000 permutations were used to test for statistical significance in both the mean performance change across the group of participants and the mean performance change within participants. 
The null hypothesis tested was that the performance of the participants using the exoskeleton with the baseline stiffness has the same distribution as the performance of the participants using the exoskeleton with the optimised stiffness. The median and the interquartile range (IQR) of the two distributions are also provided in order to identify any outliers based on the 1.5IQR rule. To quantify the variability in performance among individuals and within individuals the standard deviation and the coefficient of variation (CV) was used. Similarly, two-sided permutation tests with 100,000 permutations were used to check whether the mean behaviour of the participants when using the exoskeleton in transparent mode changed significantly beyond the training phase. A statistical analysis was not carried for the simulation results, since the simulations are deterministic and a variable number of repetitions can be carried out in simulation, which will affect variance and the results of statistical tests.

\section{Results}
\subsection{Simulation results} \label{sec:sim_results}
\begin{figure}[hb]
    \centering
    \includegraphics[width=0.99\linewidth]{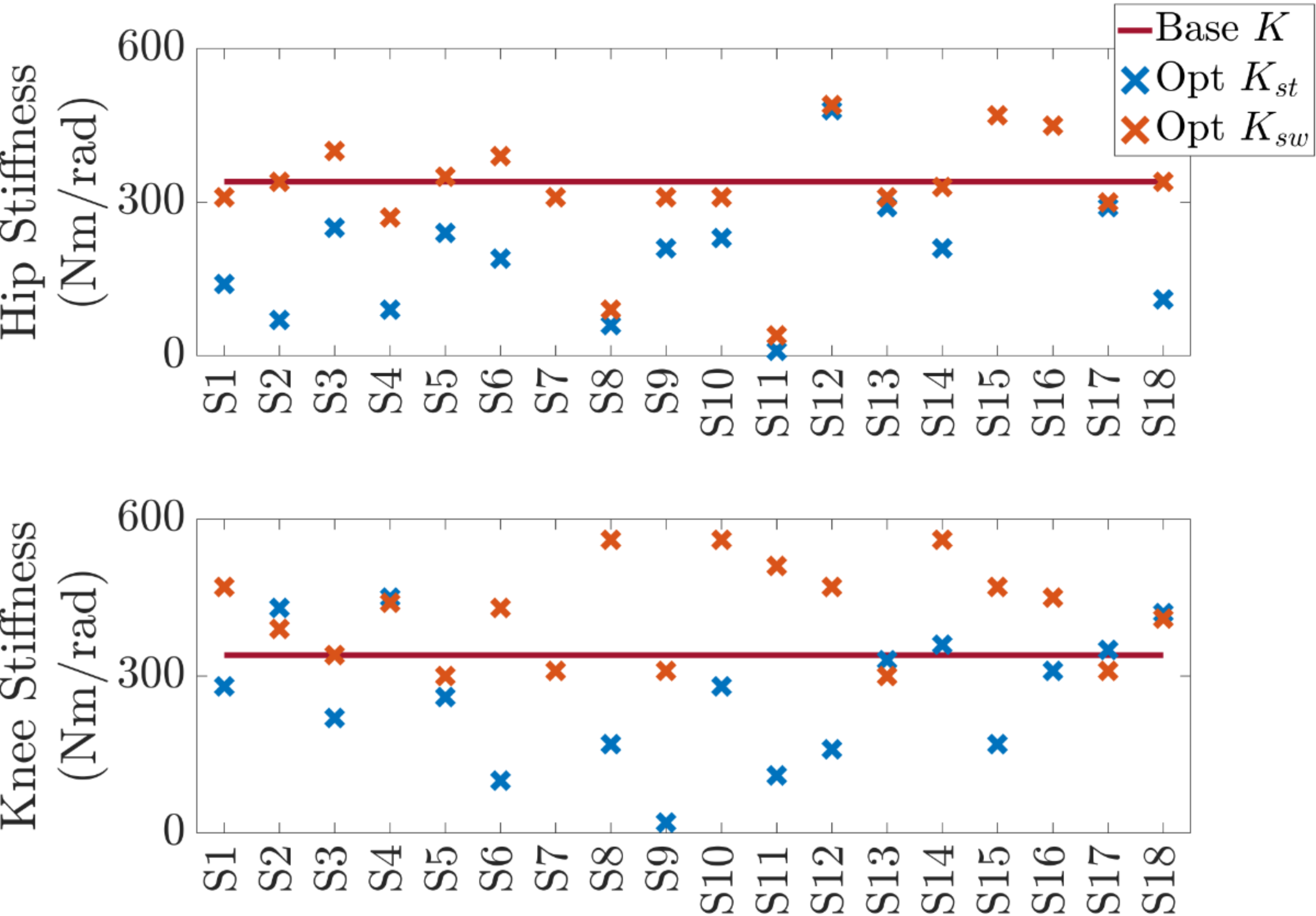}
    \caption{Personalised stiffness obtained for each participant for the two phases of the cycle for both the hip joint and the knee joint of the exoskeleton.}
\label{fig:results_stiffnesses}
\end{figure}

Figure \ref{fig:results_stiffnesses} shows the resultant stiffness obtained for each participant from the offline optimisation process for the two phases of the cycle. It can be seen that a wide range of stiffness outputs were obtained for the different participants, ranging from 20 Nm/rad (almost no assistance), to 560 Nm/rad, which corresponds to a very stiff exoskeleton. It can be observed that, for the given task, the stiffness for both the hip joint and the knee joint, for the \stancePhase, are almost always lower than the stiffness of the two joints during the \swingPhase. 

\begin{figure*}[h]
 \centering
 \includegraphics[width=0.96\textwidth]{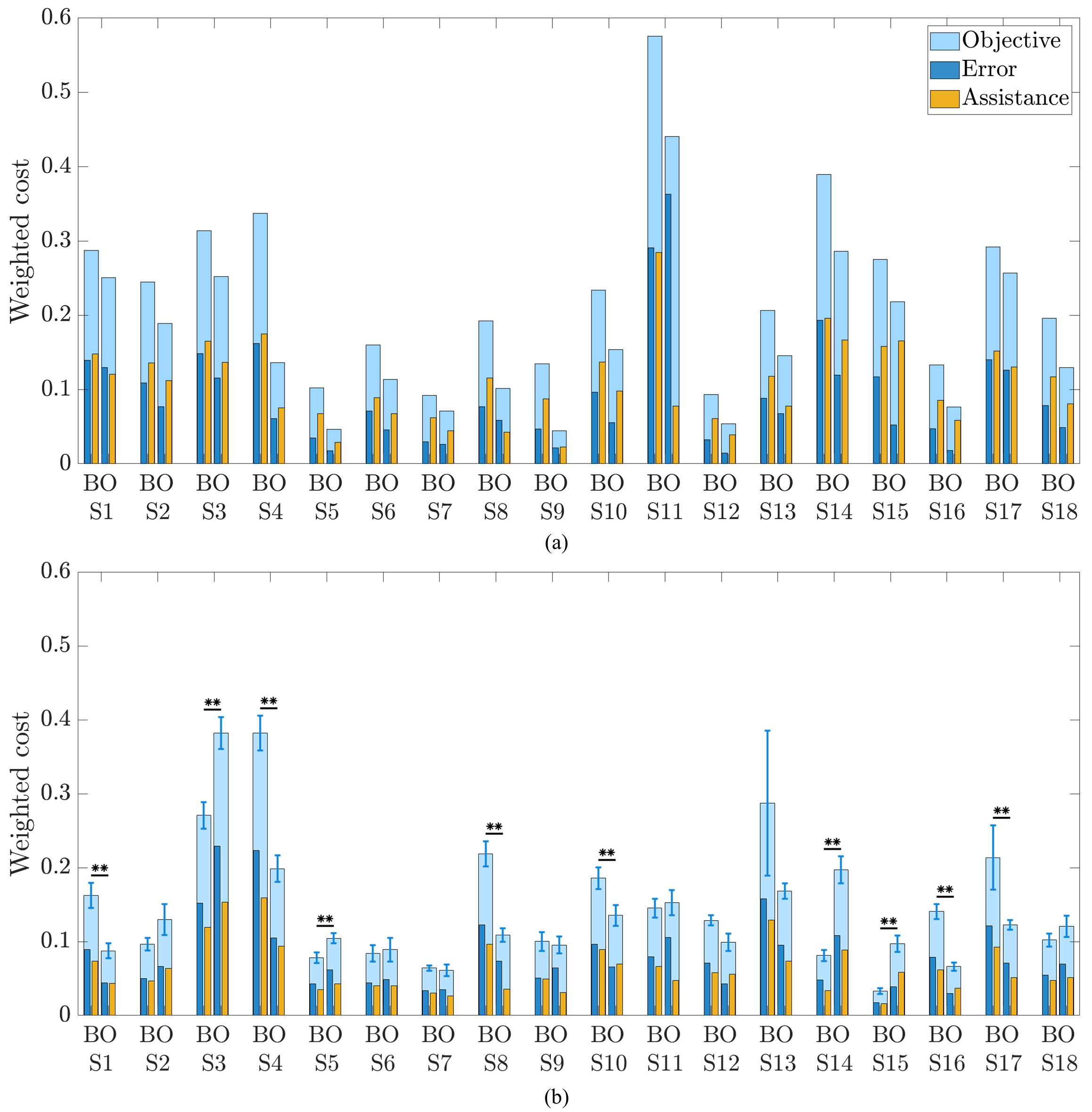} 
 \caption{(a) Simulation results, (b) Experimental results obtained when assistance from the exoskeleton was provided using the baseline controller, B, and the optimised controller, O, for the total number of cycles performed during testing. In light blue, the objective function value is shown as calculated by equation \ref{eq:ObjFun}, and in dark blue and yellow the weighted value of error and assistance are shown, respectively. (Error bars denote standard error. Statistical significance denoted with asterisks *P$<$0.05, **P$<$0.01).}
    \label{fig:results}
\end{figure*}

\begin{figure}[h]
    \centering
    \includegraphics[width=0.49\textwidth]{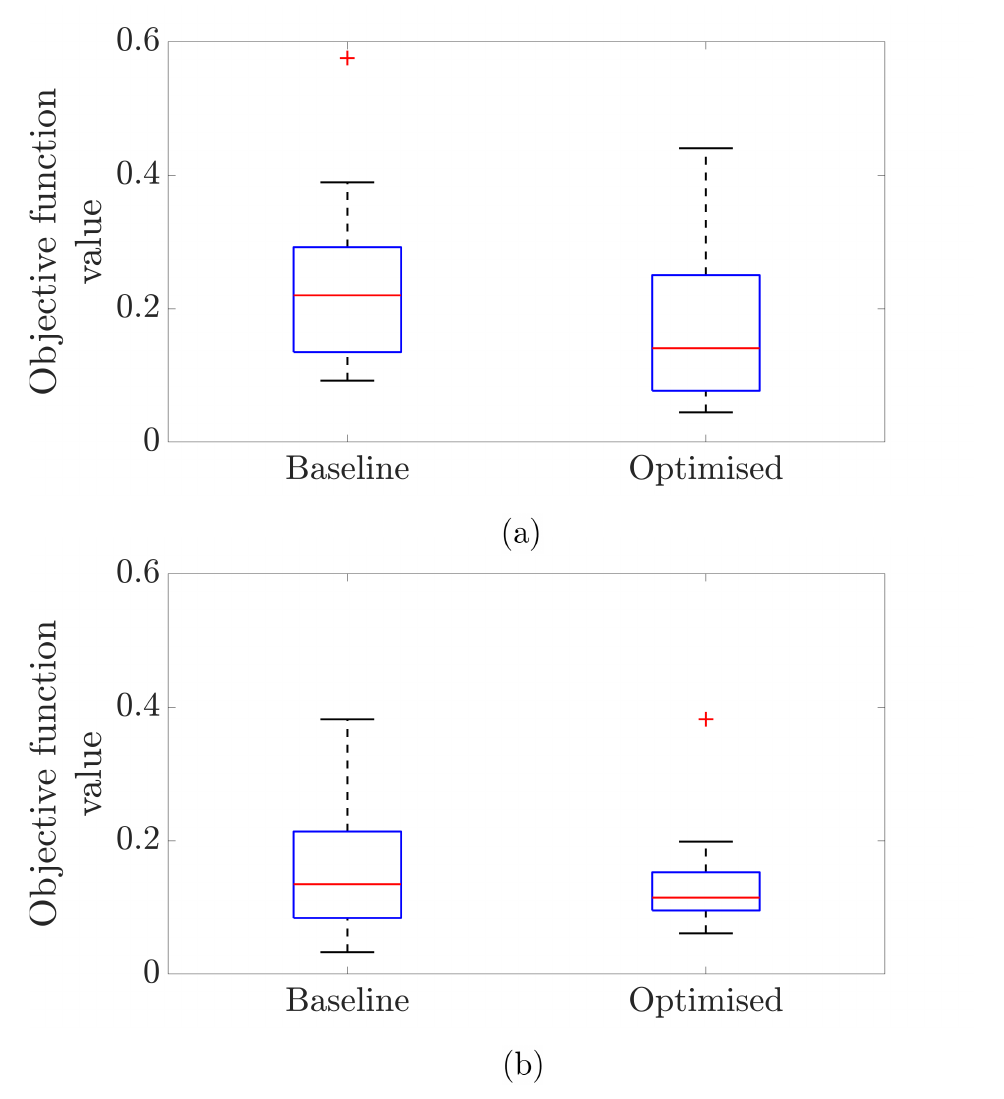}
    \caption{(a) Distribution of the expected objective function value as expressed by equation \ref{eq:ObjFun} (simulation results), (b) Distribution of the experimentally measured objective function value (experimental results).}
    \label{fig:results_means}
\end{figure}

Similarly, Figure \ref{fig:results}a shows the kinematic error and exoskeleton assistance that correspond to each participant, as predicted in simulation \footnote{Error bars on Figure \ref{fig:results}a are not provided since the simulations are deterministic and significance testing is not appropriate as a variable number of repetitions can be carried out in simulation which will affect variance and the results of statistical tests.}. It can be seen that in all cases, the offline optimisation can find an exoskeleton stiffness for the hip and the knee joints that can reduce the weighted sum of the overall tracking error of the model and the assistance provided by the exoskeleton. This consistent improvement that is predicted for the objective function value, predicts an average improvement of approximately 30.4\% (Figure \ref{fig:results_means}a) in the controller's ability to provide assistance as needed.

With the exception of S11, the obtained exoskeleton stiffness is expected to reduce both the tracking error and the levels of assistance provided. For S11, a very low stiffness was obtained for the hip joint (Figure \ref{fig:results_stiffnesses}), for both phases of the cycle, with a relatively high stiffness of the knee during the swing phase. This is expected to result in a slightly higher tracking error but significantly lower assistive forces from the exoskeleton (Figure \ref{fig:results}a). With respect to the expected performance of the rest of the participants, the expected performance of S11 using the baseline stiffness is considered an outlier based on the 1.5 IQR value (Figure \ref{fig:results_means}a). However, when the optimised stiffness is used, the expected performance of S11 lies within the 1.5 IQR value and is no longer considered an outlier (Figure \ref{fig:results_means}b). 

\subsection{Experimental results}
\subsubsection*{Task performance}
Once the optimised exoskeleton stiffness was obtained for each participant, the participants tested the exoskeleton's controller with both the baseline stiffness and the optimised stiffness in a randomised order. Their ability to follow the desired kinematic path and the levels of assistance they received from the exoskeleton are presented in Figure \ref{fig:results}b. In contrast to the simulations, during the experimental validation of the proposed method for control personalisation, not all participants performed better with the optimised exoskeleton stiffness. While for some participants the optimised exoskeleton stiffness resulted in a significant improvement in performance, as recorded by their performance per cycle, such as S1, S4, S8, S10, S16, and S17, for some participants it had no significant effect (S2, S6, S7, S9, S11, S12, S13, and S18), whereas for some it resulted in significantly worse performance (S3, S5, S14 and S15). This led to a not statistically significant change in the performance of the 18 subjects.

Throughout the experiment, the participants sometimes experienced difficulty in completing some cycles. These events led to an increased and sustained trajectory tracking error and increased assistance from the exoskeleton. This was in turn reflected in their performance as an increase in the recorded objective function value. The frequency of these events was recorded by calculating any outliers in the performance of the participants as recorded per cycle. 18 such events were observed when participants used the baseline controller, and 9 such events were observed when participants used the optimised controller. Even though cycles where such an event occurred may seem as outliers, it is unclear whether these events are independent from the choice of exoskeleton stiffness. These events were therefore not excluded from the analysis. 

It is also interesting to note that for some participants, the optimised stiffness that they experienced was very similar to the baseline stiffness of the exoskeleton, yet their performance was significantly different (S3, S5). In fact, the performance of S3 with the optimised stiffness was worse than the performance of all other participants, which may be considered an outlier according to the 1.5 IQR value (Figure \ref{fig:results_means}b). This suggests that the performance of each participant may not be entirely dependent on the stiffness of the exoskeleton controller, but may be affected by other exogenous parameters too. These may include interpersonal motor learning variability, or the participant's levels of concentration and fatigue. 

\begin{figure*}[h]
    \centering
    \includegraphics[width=0.99\textwidth]{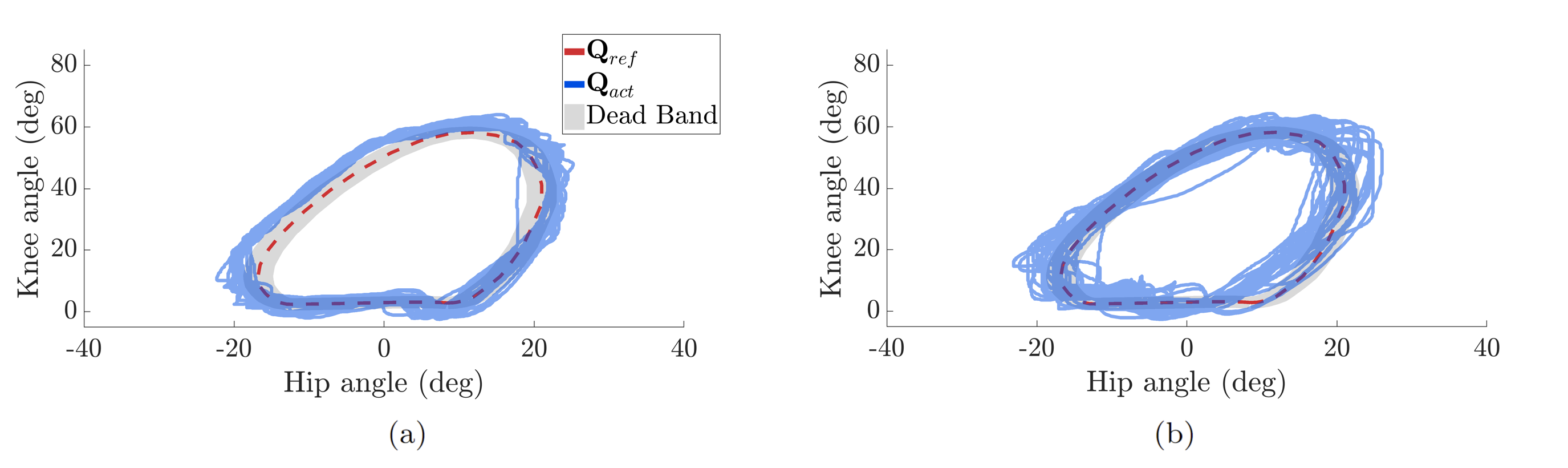}
    \caption{Recorded motion of two participants while wearing the exoskeleton in assistive mode. (a) The recorded motion of a participant who had less variable movement (S5), and (b) the recorded motion of a participant who had more variable movement (S11).}
    \label{fig:intra-personal variability}
\end{figure*}

To quantify this variability in performance among individuals, and within individuals, the standard deviation and the coefficient of variation (CV) was calculated ($CV=\sigma/\mu$) for their performance during the assistive trials. A standard deviation of 0.091 and 0.073 and a CV of 0.59 and 0.55 was obtained for the performance of the participants while using the baseline controller and the optimised controller, respectively. This indicates a high inter-personal variability, where the standard deviation is often more than half the average performance of the participants. Similarly, the standard deviation and the CV were calculated for the performance of each participant independently and as it was recorded per cycle during the assistive trials. The standard deviation of the participants' performance and the corresponding CV using the two controllers ranged from 0.015-0.480 and 0.23-1.67, respectively. This indicates that some participants performed more consistently than others, while some participants had significant variations in their motor commands during the execution of the task, which were independent of the controller used. For reference, Figure \ref{fig:intra-personal variability} shows the performance of two individuals with similar performance, as quantified by their ability to follow the reference path, but with different variability in performance. When this variation in performance is compared to the expected 30\% improvement due to the personalisation of the controller parameters (section \ref{sec:sim_results}), a low signal-to-noise ratio can be noticed.

\subsubsection*{Task learning variability}
\begin{figure}[h]
    \centering
    \includegraphics[width=0.99\linewidth]{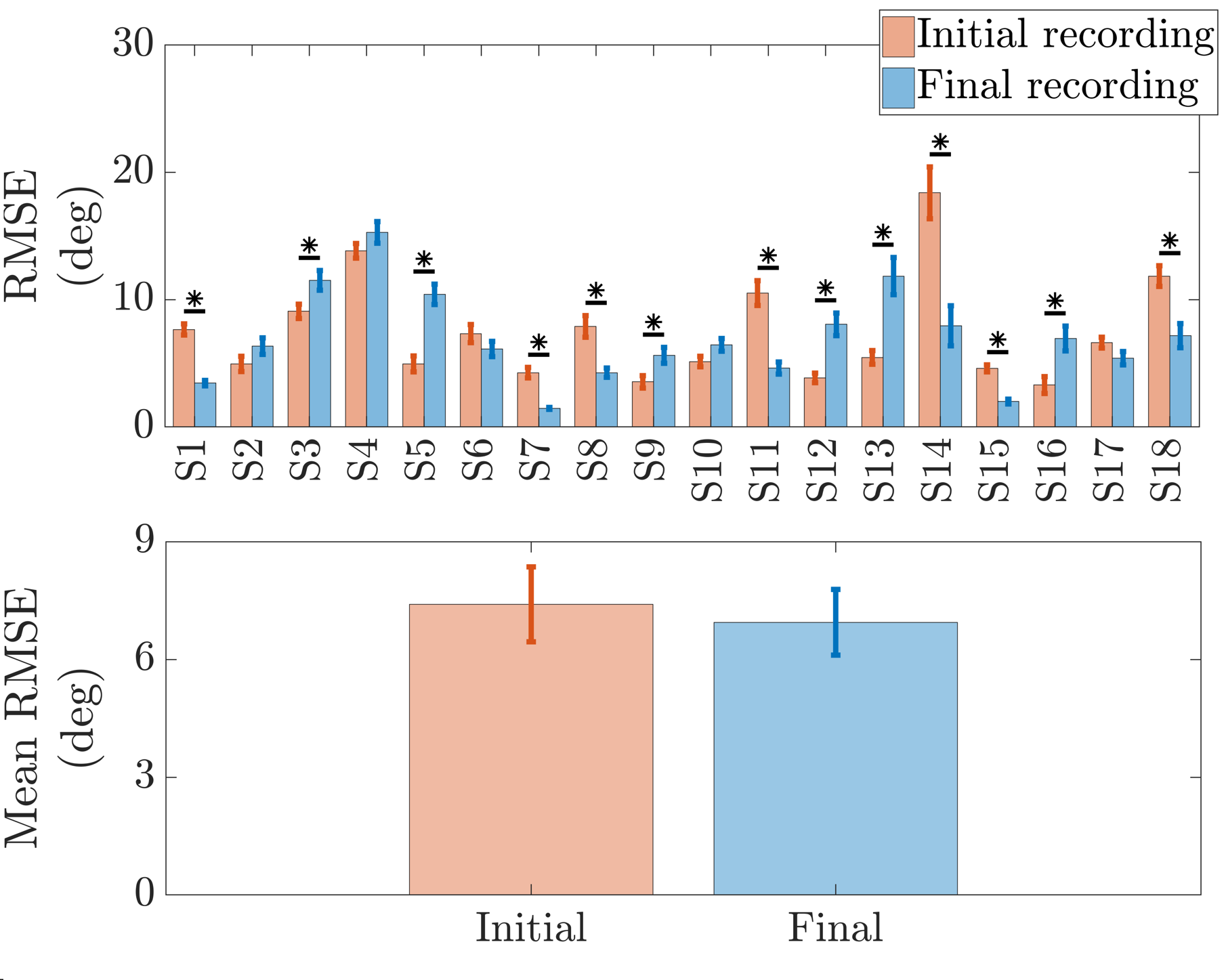}
    \caption{Kinematic tracking error of participants at the beginning of the experiment and at the end of the experiment. (Error bars denote standard error. Statistical significance denoted with an asterisk for P$<$0.05.)}\label{fig:rmse_before_and_after}
\end{figure}

After the participants experienced the exoskeleton's assistive controllers, they were asked to perform the task with the exoskeleton in transparent mode once again. This was to investigate any behaviour changes that may have occurred during the experiment, and test the validity of our assumption that the behaviour of the participants will not significantly change beyond the training phase. The results can be seen in Figure \ref{fig:rmse_before_and_after}. While the performance of some individuals was observed to change significantly, the change in the overall performance of the participants as a group was not statistically significant. A high variance is also noticeable, reflecting the fact that some participants were able to track the reference path more accurately than others. This again may be a result of external factors such as concentration, fatigue and potentially even some motor learning, affecting the behaviour of the participants. 

\begin{figure*}[h]
    \centering
    \includegraphics[width=0.99\textwidth]{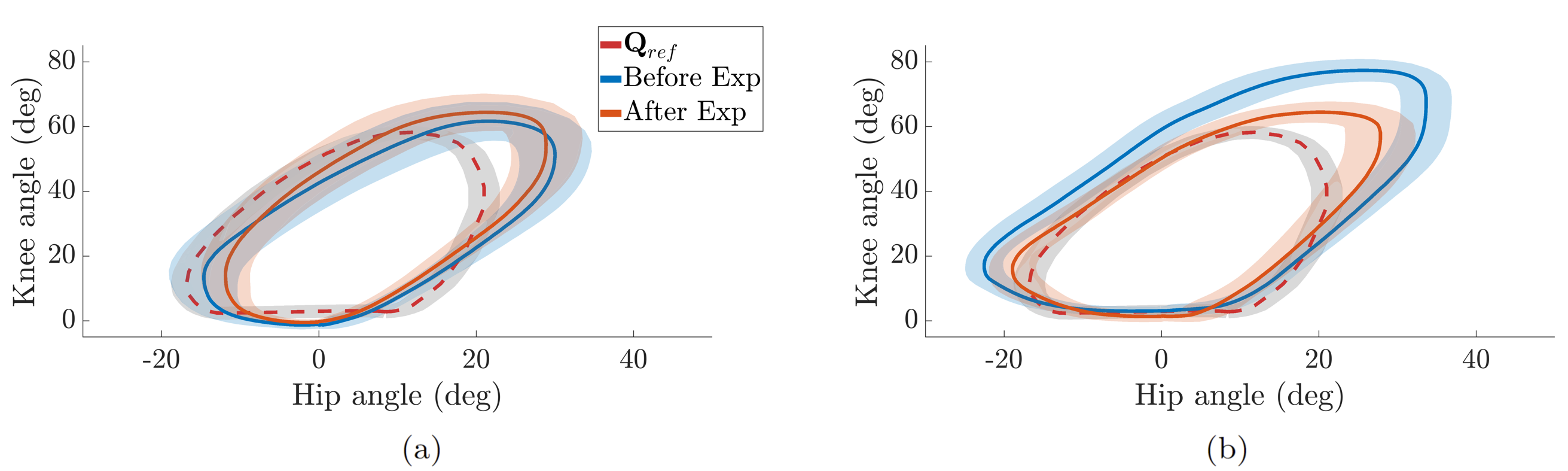}
    \caption{Recorded motion of two participants before and after the experiment while wearing the exoskeleton in transparent mode. (a) The recorded motion of a participant who had no significant improvement in their performance (S2), (b) The recorded motion of a participant who had a significant improvement in their performance (S11).}
    \label{fig:rmse_before_and_after_S2_S11}
\end{figure*}

For reference, Figure \ref{fig:rmse_before_and_after_S2_S11} illustrates the recorded motion of two participants before and after the experiment; one participant who had no evident change in their motor commands (S2), and one participant who had a significant change in their motor commands throughout the experiment (S11). It can be seen that while initially both participants demonstrated an exaggerated hip flexion and knee flexion during the \swingPhase, after approximately 5 minutes of testing, subject S2 had no obvious change in their kinematics, while subject S11 had a significant decrease in their range of motion during the \swingPhase, which more accurately follows the reference path.

Figure \ref{fig:rmse_before_and_after_S2_S11} can also be used to make a comparison between the participants' unassisted kinematics and their assisted kinematics (Figure \ref{fig:intra-personal variability}). It can be seen that due to the assistive forces provided by the robot, the tracking performance of the participants improves. This suggests that the robot's controller is able to reliably correct the participants' deviations from the reference path.

\section{Discussion}
Here, we presented an offline model-based approach for the design and personalisation of robotic controllers that focus on human-robot collaboration. With the help of eighteen healthy participants, we demonstrated how this approach can be applied to the tuning of an AAN controller using a lower-limb exoskeleton. The results indicate that the personalisation of robotic controllers using offline optimisation may be able to improve the collaboration between the user and the robot, particularly as our understanding of the dynamic interaction between human and robot improves. In simulation, it is clear that a unique exoskeleton stiffness can be obtained for each participant, which is expected to improve the participant's ability to follow the desired path more accurately and with less robotic assistance. This implies an improved ability to provide assistance as needed and promote patient-driven rehabilitation. During our experimental validation, this improvement was not uniform across all the participants. Inter- and intra-personal variability was evident. Performance variability, especially among individuals whose personalised stiffness was similar to the baseline stiffness, suggests that other factors such as concentration, fatigue and motor learning ability may be influencing the results. It is possible that modelling uncertainty and low signal-to-noise ratio have detracted from the benefits observed experimentally. 

One of the main challenges that may have contributed to modelling uncertainty, is the modelling of human behaviour. In this study, a feedforward model of human behaviour was constructed based on the recorded motion of the participants. A short sample of five cycles was selected from the recorded motion in order to reduce computational demands, and it was assumed that human behaviour can be accurately captured within the cycles selected from the recorded data. However, it is likely that movement variability and the participants' response to assistive controllers of varying stiffness have challenged this assumption. The participants’ ability to perceive the exoskeleton's assistance and the change in the exoskeleton's controls may have biased their response in order to satisfy an objective function that may be different from the one expressed by equation \ref{eq:ObjFun}, e.g. an objective function that also aims to minimise effort and increase comfort. It is expected that with a better estimation of the human behaviour and movement variability, the improvements observed in simulation will more closely match the observed improvements in real life. Future studies can consider increasing the duration of the training period, as this may decrease movement variability, and focus on developing more accurate human models that can more reliably capture movement variability while accounting for the human tendency to minimise effort.

In \cite{Gordon2023} the authors also proposed a method for learning a personalised human behaviour model for the execution of a sit-to-stand task while wearing a lower-limb exoskeleton. This approach uses bi-level optimisation and is based on the assumption that the human behaviour can be characterised by a neural multi-objective optimisation process, where the objective comprises fundamental principles such as balance, energy consumption, and joint loading. Such a model could provide an estimation of the human adaptation to external forces but the computational demands for obtaining a personalised human behaviour model are high, and the fundamental principles that need to be considered for the composition of an accurate objective function for different tasks can be hard to identify. Computational models of human motor learning are also proposed in \cite{Tee2010,Scheidt2001,Thoroudhman2000} based on the ability of healthy people to adapt their strategy to achieve a motor task in environments where perturbations are provided in the form of unpredictable perturbations or predictable force fields with or without stochastic catch trials. However, it is unknown whether these models can capture the adaptation of human behaviour in the presence of a predictable force field in a collaborative task. 

Emken et al. \cite{Emken2007c,Emken2005} used these computational models to describe the adaptive behaviour of unimpaired participants when subjected to a virtual impairment in the form of a robotic force field. Using an objective function that included an error cost and an assistive cost, Emken et al., proceeded to analytically derive a robotic controller that provides assistance as needed. This led to an error-based adaptive controller that bounds kinematic errors and reduces its assistance as the performance of the user improves \cite{Emken2007c}. This is different to the presented framework where the emphasis is on optimising such controllers to the needs of the user. In fact the approach adopted by Emken et al., provides another example of where the presented framework could be useful to facilitate the personalisation of the open parameters of the adaptive controller with the help of personalised musculoskeletal models. 

When trying to bridge the gap between simulation and reality, a better understanding of the contact dynamics between human and robot is also required. In many studies, fixed contact points have been used \cite{Khamar2019,Ren2019}, but this approach does not take into account force transmission losses and may result in restricted motion in the presence of joint misalignment. Kinematic constraints have also been used \cite{Inkol2020,Torricelli2018c,Chen2018c}. This method avoids constraints that may arise due to joint misalignment but may rely in unrealistic interaction forces between the human and the robot. In this study, bushing forces have been used at the interaction points between the human model and the exoskeleton, which are a combination of translational and rotational elastic and viscous forces. This allows for the use of plausible human-exoskeleton interaction models, where factors such as transmission losses and skin elasticity can be considered \cite{Hosseini-zahraei2022,Duong2016}. However, an accurate value of the stiffness and the damping coefficient of these bushing forces is hard to obtain. In \cite{Serrancoli2019}, Serrancoli et al. proposed a method for estimating human-exoskeleton contact forces, as well as ground contact forces in sit-to-stand movements. Such calibration routines could improve the accuracy of the human-exoskeleton models and the effectiveness of offline model-based optimisation.

Lastly, it is prudent to consider the usability and acceptance of the proposed pipeline by all stakeholders. On the one hand, compared to conventional robotic interventions that are not personalised to the needs of the user, the proposed pipeline may be able to lead to improved outcomes, but at the cost of a slightly increased workload for both the operator and the patient, in order to personalise the robotic controller to each patient. Given the already overloaded schedule of healthcare providers, it is important to consider methods for improving modelling accuracy through computationally efficient algorithms, to then be able to investigate the usability and acceptance of model-based optimisation through clinical experiments. On the other hand, compared to other personalisation methods such as adaptive control or human-in-the-loop optimisation, which often require the patient's active participation during the exploration and exploitation phases, the proposed pipeline is expected to improve usability for patients by eliminating their involvement during these stages, thereby reducing their cognitive and physical burden.

\section{Conclusions}
In this paper we presented a framework for the design and fine-tuning of personalised robotic controllers utilising high-fidelity musculoskeletal models. The proposed method offers a means to reduce the reliance on extensive human-in-the-loop testing and improve the collaboration between human and robot in order to increase productivity, comfort, safety, and learning. Illustrated through a case study focusing on a collaborative lower-limb task, we demonstrated the feasibility and potential benefits of employing offline model-based optimisation. We observed that in an ideal environment the right tuning of a robotic controller can have a significant impact on the ability of the robot to support the user as needed. In practice though, exogenous effects such as concentration, fatigue and interpersonal and intra-personal variability in motor control and motor learning can partly inhibit the expected improvements. This further highlights the importance of interventions that will more accurately capture these interpersonal and intra-personal variations. This calls for future studies that will contribute towards our understanding of human motor control and learning in collaborative tasks with robots, and the expansion of methodologies that facilitate the personalisation of assistive robots. The present study thus constitutes a promising proof of concept laying the foundation for further exploration into model-based optimisation for the design and personalised tuning of controls for assistive and collaborative robots.

\section*{Acknowledgments}
Thanks to the Neural Rehabilitation Group at Cajal Institute, Madrid, Spain, and the SLMC group, at the University of Edinburgh, UK, for taking part in the experiments and their constructive feedback.

\bibliographystyle{IEEEtran}

\vspace{1pt}

\begin{IEEEbiography}[{\includegraphics[width=1in,height=1.25in,clip,keepaspectratio]{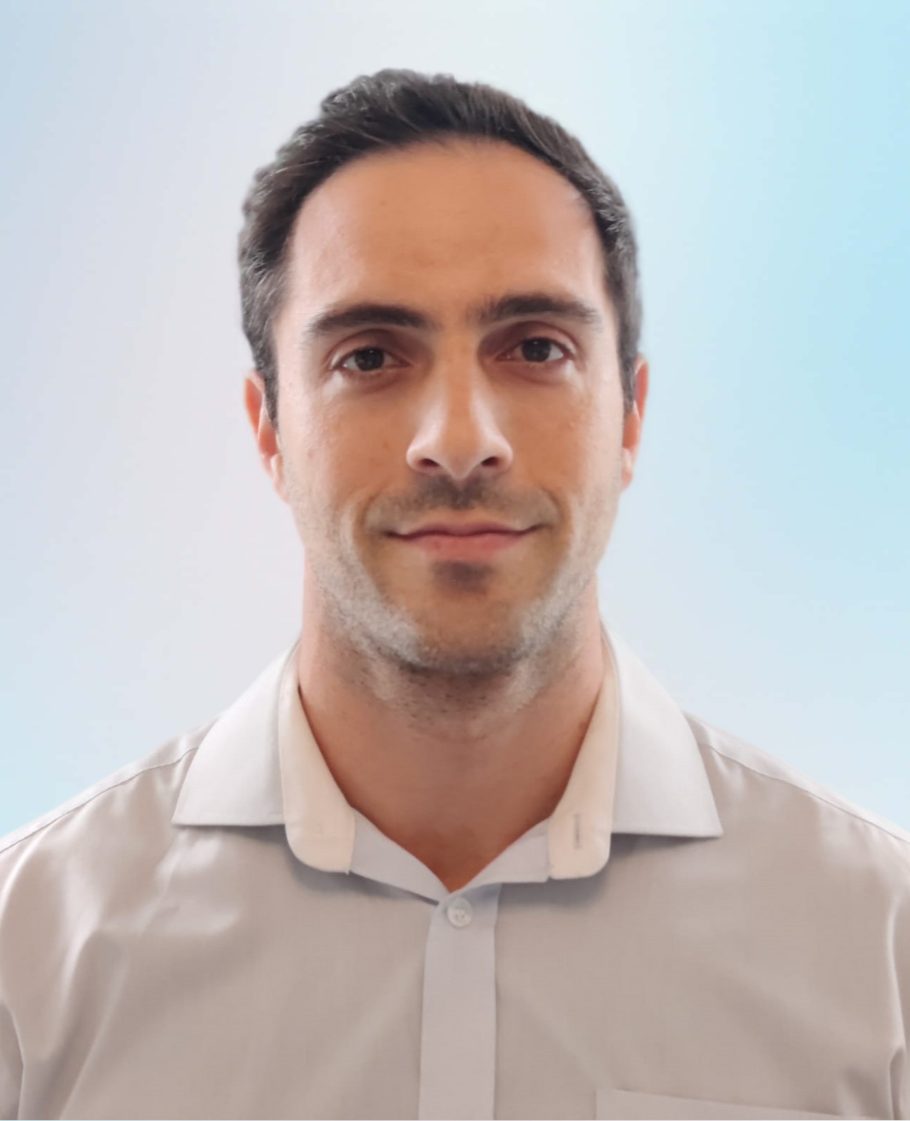}}]{Andreas Christou}
is a Research Associate at the Statistical Machine Learning and Motor Control ~(SLMC) Group at the University of Edinburgh, UK. He received the Master of Engineering (M.Eng.) degree in Mechanical Engineering in 2019 from the University of Edinburgh, UK, where he also participated in the Education Abroad Programme at the University of California, Santa Barbara. In 2024, Andreas received his Ph.D. in Robotics and Autonomous Systems at the Centre for Doctoral Training at the University of Edinburgh, UK. His research interests include the use of wearable robots and functional electrical stimulation for gait rehabilitation and human augmentation. 
\end{IEEEbiography}

\begin{IEEEbiography}[{\includegraphics[width=1in,height=1.25in,clip,keepaspectratio]{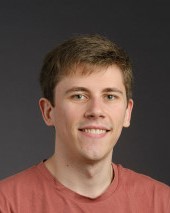}}]{Daniel F. N. Gordon}
is a Robotics Research Engineer at Huawei's German Research Center in Munich, Germany. He received the M.Math. degree in mathematics from the University of St Andrews, UK, in 2014, the M.Sc.(R) degree in robotics and autonomous systems from the University of Edinburgh, UK, in 2016 and the Ph.D. degree in exoskeleton-assisted locomotion from the University of Edinburgh, UK, in 2021. He was a Research Associate with the University of Edinburgh until 2023, exploring topics related to gait analysis, musculoskeletal modelling, optimisation, and the control of assistive robotic devices.
\end{IEEEbiography}


\begin{IEEEbiography}[{\includegraphics[width=1in,height=1.25in,clip,keepaspectratio]{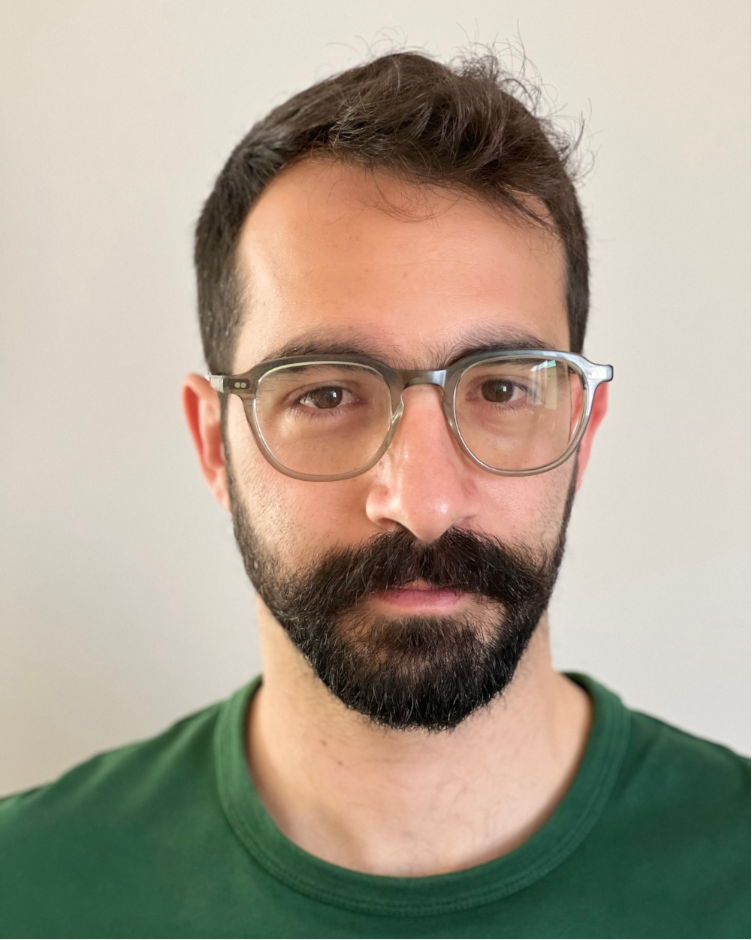}}]{Theodoros Stouraitis}
is a Control Scientist working on robotics and autonomous vehicles at DeepSea Technologies, Greece. He received his Ph.D. in Robotics from the University of Edinburgh, UK, in collaboration with the Honda Research Institute Europe in Germany. From 2012 to 2015, he was a research assistant at the Institute of Robotics and Mechatronics of the German Aerospace Center (DLR). In 2022, he was a Visiting Scientist at Massachusetts Institute of Technology (MIT), USA, and a Guest Scientist at the Honda Research Institute Europe, Germany. His research interests include motion planning and control, non-linear optimisation, contacts, and human-robot interaction.

\end{IEEEbiography}


\begin{IEEEbiography}[{\includegraphics[width=1in,height=1.25in,clip,keepaspectratio]{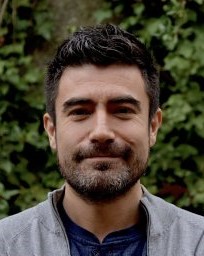}}]{Juan C. Moreno}
is Principal Investigator and Director of the BioRobotics Group in Center for Automation and Robotics at CSIC (Spain). He is the Responsible of the Associated Unit of CSIC for the National Spinal Cord Injury Hospital promoting translational research to generate knowledge and deliver solutions for disabled people. One of his main lines of research investigates the benefits of cooperative control in the fusion of neuro-prosthetics and robotic exoskeletons in improving locomotion in stroke and patients with spinal cord injury. He is currently leading research projects on modelling and characterisation of longitudinal robot-aided treatments towards optimisation of robotic- and FES-based therapy with intelligent controllers. Dr Moreno is the founder \& Co-Chair of the IEEE/RAS Technical Committee (TC) on Wearable Robotics and a co-founder of the company Technaid S.L., Spain, a `spin-off' company of CSIC.
\end{IEEEbiography}


\begin{IEEEbiography}[{\includegraphics[width=1in,height=1.25in,clip,keepaspectratio]{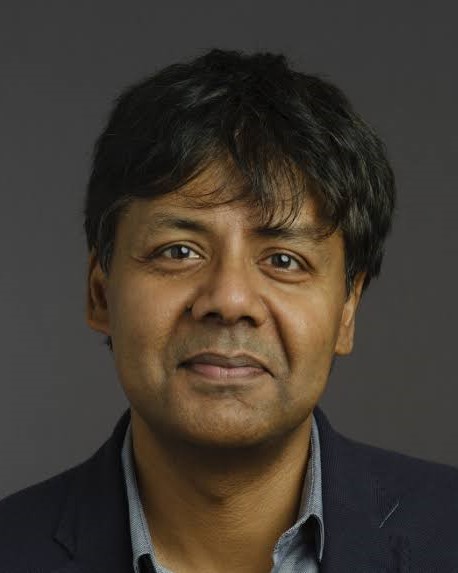}}]{Sethu Vijayakumar}
received the Ph.D. degree in computer science and engineering from the Tokyo Institute of Technology, Tokyo, Japan, in 1998. He is the Professor of Robotics with the University of Edinburgh, an adjunct faculty of the University of Southern California, Los Angeles, and the founding Director of the Edinburgh Centre for Robotics. His research interests include statistical machine learning, 
anthropomorphic robotics, planning, multi-objective optimisation, and optimal control in autonomous systems as well as the study of human motor control. Prof. Vijayakumar helps shape and drive the national Robotics and Autonomous Systems agenda in his recent role as the Programme Co-Director for artiﬁcial intelligence with the Alan Turing Institute, U.K.’s national institute for the Data Science and AI. He is a Fellow of the Royal Society of Edinburgh, and winner of the 2015 Tam Dalyell Prize for excellence in engaging the public with science.
\end{IEEEbiography}


\vfill

\end{document}